\documentclass[conference]{IEEEtran}

\usepackage{cite}
\usepackage{amsmath,amssymb,amsfonts}
\usepackage{algorithmic}
\usepackage{graphicx}
\usepackage{textcomp}
\usepackage{xcolor}
\PassOptionsToPackage{bookmarks=false}{hyperref}
\PassOptionsToPackage{hyphens}{url}
\usepackage{hyperref}
\def\BibTeX{{\rm B\kern-.05em{\sc i\kern-.025em b}\kern-.08em
T\kern-.1667em\lower.7ex\hbox{E}\kern-.125emX}}

\usepackage{tikz}

\newcommand\submittedtext{%
	\footnotesize This work has been submitted to the IEEE for possible publication. Copyright may be transferred without notice, after which this version on arXiv will be replaced with the Version of Record.}

\newcommand\submittednotice{%
	\begin{tikzpicture}[remember picture,overlay]
		\node[anchor=south,yshift=10pt] at (current page.south) {\fbox{\parbox{\dimexpr0.65\textwidth-\fboxsep-\fboxrule\relax}{\submittedtext}}};
	\end{tikzpicture}%
}

\begin{document}

    \title{Keeping it Local, Tiny and Real:\\ Automated Report Generation on Edge Computing Devices for Mechatronic-Based Cognitive Systems\\
    }
    \author{
        \IEEEauthorblockN{\begin{tabular}{ccc}
                              1\textsuperscript{st} Nicolas Schuler\IEEEauthorrefmark{1}\IEEEauthorrefmark{2} & 2\textsuperscript{nd} Lea Dewald\IEEEauthorrefmark{1} & 3\textsuperscript{rd} Jürgen Graf\IEEEauthorrefmark{1}\\
                              \href{https://orcid.org/0009-0007-4098-1244}{{\includegraphics[keepaspectratio,width=0.7em]{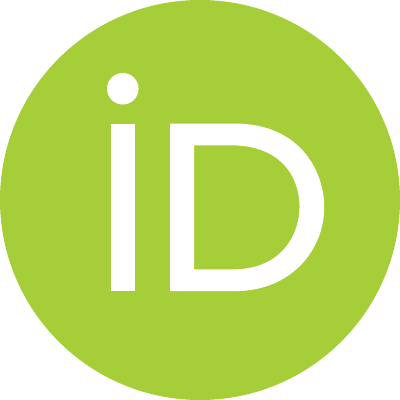}}} 0009-0007-4098-1244 &
                              \href{https://orcid.org/0009-0004-8825-0545}{{\includegraphics[keepaspectratio,width=0.7em]{images/ORCID_iD.pdf}}} 0009-0004-8825-0545 & \href{https://orcid.org/0000-0002-1354-0888}{{\includegraphics[keepaspectratio,width=0.7em]{images/ORCID_iD.pdf}}} 0000-0002-1354-0888
        \end{tabular}}
        \IEEEauthorblockA{\begin{tabular}{cc}
                              \\ \IEEEauthorrefmark{1} \textit{Department of Computer Science} & \IEEEauthorrefmark{2} \textit{Department of Engineering}\\
                              \textit{Trier University of Applied Sciences} & \textit{University of Luxembourg}\\
                              Trier, Germany & Kirchberg, Luxembourg\\
                              \{schulern,lxdw0338,j.graf\}@hochschule-trier.de & nicolas.schuler.001@student.uni.lu
        \end{tabular}
        }
    }

    \maketitle

    \begin{abstract}
        Recent advancements in Deep Learning enable hardware-based cognitive
        systems, that is, mechatronic systems in general and robotics in
        particular with integrated Artificial Intelligence, to interact with
        dynamic and unstructured environments.
        While the results are impressive, the application of such systems to
        critical tasks like autonomous driving as well as service and care
        robotics necessitate the evaluation of large amount of heterogeneous
        data.
        Automated report generation for Mobile Robotics can play a crucial
        role in facilitating the evaluation and acceptance of such systems in
        various domains.
        In this paper, we propose a pipeline for generating automated reports
        in natural language utilizing various multi-modal sensors that solely
        relies on local models capable of being deployed on edge computing
        devices, thus preserving the privacy of all actors involved and
        eliminating the need for external services.
        In particular, we evaluate our implementation on a diverse dataset
        spanning multiple domains including indoor, outdoor and urban
        environments, providing quantitative as well as qualitative evaluation
        results.
        Various generated example reports and other supplementary materials are
        available via a public repository.
    \end{abstract}

    
	\submittednotice

    \section{Introduction}\label{sec:introduction}
    Recent developments in Deep Learning, especially Large Language Models (LLMs)
    and Vision-Language Models (VLMs), allow for mobile cognitive systems to
    operate in dynamic and unstructured environments, including task
    planning\cite{Chen2024}, object detection\cite{Liu2023}, understanding human
    behaviour\cite{Karim2025} and finally reasoning over the relationships between
    these\cite{Cherian2024}.
    The incorporation of various sensors systems, including camera, LiDAR and
    INS/GNSS, and manipulators allow these systems to perceive their environment
    and to interact with it\cite{Beetz2025}.
    Such cognitive systems are of particular importance in the field of autonomous
    driving\cite{Zhao2025}, service robotics and robotic
    caretakers\cite{SilveraTawil2024}, where the understanding of urban and indoor
    environments is a complex and critical tasks for human-robot interaction,
    cooperation and collaboration.

    The evaluation, supervision and monitoring of such platforms is challenging,
    considering the vast range of domains and potential environments they are
    employed in\cite{Tang2024}.
    The mobile agent might succeed in highly complex unstructured environments of
    a certain type but completely fail in another\cite{Wang2025}, with the
    difference in domain and difficulties associated to it not being obvious for a
    human observer.
    Additionally, collective and swarm robotics\cite{Feng2020} place much more
    pressure on supervision, by increasing the number of agents, interactions and
    monitoring tasks considerably.
    Finally, the often sensitive nature of the domains, i.e.\ recording of data
    subject to privacy laws, and amount of received sensory data necessitate the
    filtering of data that can be put to human revision.
    Besides the challenges associated with the amount and type of data, the
    data interpretation itself might be challenging, requiring specialists to
    evaluate or make the data available for further computations to begin with.

    In this paper, we introduce a pipeline for generating automated reports for
    mobile agents operating in arbitrary domains that may be utilized for
    multiple tasks including evaluation and map generation.
    We do this locally, on an edge computing device and with models capable of
    real time inference.
    To that end, we incorporate various Deep Learning architectures that are
    not fine-tuned to our particular domains to generate textual descriptions
    of the perceived scenes and actions therein, detect and segment objects and
    agents of interest to these descriptions, localize these detections within
    world coordinates, group these actions by similarity and anonymize the
    output to preserve the privacy of all agents actively and passively
    involved into the process.
    This results in automatically generated documents that can be understood by
    a layman who might not be knowledgeable about the details of the particular
    Deep Learning architectures utilized.
    Finally, we achieve this utilizing edge computing devices, ensuring that the
    data recorded is only processed locally either directly on the edge device or,
    depending on the requirements on accuracy and computational budgets, on local
    devices, completely removing the usage of cloud-based solutions and larger,
    external models that would not be under our direct control and data governance.
    Supplementary materials are available via a public
    repository,\footnote{\hspace{1 pt} \url{https://datahub.rz.rptu.de/hstr-csrl-public/publications/automated-report-generation-robotics/}}
    including code, high resolution image, video examples, evaluation data and
    fully generated reports.
	While not limited to, this work focuses on mechatronic-based cognitive systems
	acting in real-world scenarios.

    The remainder of this paper is structured as follows:
    Sect.~\ref{sec:foundations-and-related-work} gives an overview of various
    models important to our pipeline and related work in automated, Deep Learning
    supported report generation with a focus on robotics.
    Sect.~\ref{sec:automated-report-generation-pipeline} introduces the proposed
    pipeline and the usage within our laboratory, with the pipeline being evaluated
    and discussed in Sect.~\ref{sec:experimental-evaluation}.
    Finally, Sect.~\ref{sec:conclusion-and-future-work} gives a short summary of
    the findings and future work.

    \section{Foundations and Related Work}\label{sec:foundations-and-related-work}

    The following section introduces various Deep Learning architectures and
    models that can be applied to the task of natural language report generation
    with a focus on so-called foundation models, large pre-trained machine learning
    models capable of being applied to a large variety of tasks in the field or
    robotics\cite{Xiao2025}.
    First, Sect.~\ref{subsec:vision-language-models} gives a short introduction to
    LLMs and VLMs, which can be
    utilized for various reasoning and language generation tasks.
    Following, Sect.~\ref{subsec:zero-shot-detection-and-segmentation-foundation-models}
    focuses on zero-shot object detection and segmentation models, that is models
    which can be used to detect arbitrary objects without the need for fine-tuning
    to a specific domain.
    Finally, Sect.~\ref{subsec:related-work} discusses the related work of
    automated report generation using such foundational models, focusing on their
    application in autonomous driving and robotics.

    \subsection{Vision Language Models}\label{subsec:vision-language-models}

    The recent introduction of the transformer-based architectures\cite{Vaswani2017}
    of LLMs like ChatGPT\cite{OpenAI2023} and VLMs like Gemini Robotics\cite{GRT2025}
    allows mobile platforms to gain a deeper knowledge about their environment than
    what is possible with pure object detection, leading to more sophisticated
    interactions with that environment.
    Beyond the pure text-based interactions with LMMs, VLMs utilize multi-modal
    data, that is image data in combination with text, and thus can be used to
    generate descriptions of individual images or even entire sequences and reason
    over them\cite{Zellers2018}, including implicit knowledge not immediately
    available within the provided images.

    In particular, small Vision-Language Models (sVLMs) enable the usage of such
    models even on edge computing devices, e.g.\ Moondream2\cite{vik2024},
    FastVLM\cite{Vasu2024} and SmolVLM2\cite{Marafioti2025}.
    While such models display strong results, their restricted size leads to
    inherent biases: for a recent overview see\cite{Patnaik2025}.

    \subsection{Zero-Shot Detection and Segmentation Models}\label{subsec:zero-shot-detection-and-segmentation-foundation-models}

    The shift towards transformer-based architectures is also present in the
    classical computer vision domains of object detection and
    segmentation\cite{Zou2023}.
    While Convolutional Neural Networks (CNNs) dominated these task for the past
    decade, recently attention-based\cite{Vaswani2017} Vision Transformers
    (ViTs)\cite{Dosovitskiy2020} or hybrid architectures came to dominate
    vision-based object detection and segmentation, especially when utilizing
    multi-modal data\cite{Carolan2024}, to capture long-range temporal and spatial
    dependencies\cite{Wang2024}, in such use cases where the general longer
    inference times and higher computational budget of transformers
    are of no immediate concern\cite{ChittyVenkata2023} and up to reasoning about
    the physics observed in the scene\cite{Cherian2024,GRT2025}.

    The zero-shot capabilities provided by transformer-based models allow to detect
    and segment arbitrary object classes without any domain-specific fine-tuning.
    Open-vocabulary tasks\cite{Yu2025} enable the prompting of such models with natural
    language texts, e.g.\ for object detection and segmentation.
    Popular models include SAM\cite{Kirillov2023} and DINOv3\cite{Simeoni2025} for
    segmentation tasks, Grounded DINO\cite{Liu2023} and OWLv2\cite{Minderer2023}
    for object detection.

    \subsection{Related Work}\label{subsec:related-work}

    Automated report generation is especially common in medical
    applications\cite{Ji2024,Rao2025}.
    Closer to our application,\cite{Pu2024} apply LLMs and SAM\cite{Kirillov2023}
    for automated construction inspection reports.
    Focusing on applications within the field of robotics,\cite{Abbas2024} combine
    LLMs and VLMs to serve as a mutual human-computer interface,
    while~\cite{Wu2025} utilize VLMs to generate simulated hazardous testing
    scenarios.
    For autonomous driving,\cite{Bai2024} apply 3D-tokenized LLMs for object
    detection, map construction and task planning.
    Concerning road incidents,\cite{Farhan2025} utilize a combination of local
    object detection on UAVs, a local VLM (Moondream2\cite{vik2024}) and the
    GPT-4\cite{OpenAI2023} Turbo API for automated incident localization,
    documentation and response.

    To the best of our knowledge, the approach presented in this paper is unique in
    that we use only small, local Deep Learning models to generate human-readable
    reports utilizing Semantic Similarity measures to aggregate scene descriptions
    while preserving full data governance and being completely open to the type of
    mechatronic-based cognitive system and domain.

    \section{Automated Report Generation Pipeline}\label{sec:automated-report-generation-pipeline}

    This section introduces the proposed pipeline to automatically generate
    reports for mobile cognitive systems.
    Sect.~\ref{subsec:mobile-platforms-and-sensory-systems} introduces the
    cognitive systems and sensors we use in our laboratory and
    Sect.~\ref{subsec:automated-report-generation-for-mobile-robotics} gives an
    overview of the proposed architecture.

    \subsection{Our Mobile Platforms and Sensory Systems}\label{subsec:mobile-platforms-and-sensory-systems}

\begin{figure}[htbp]
        \centerline{\includegraphics[width=\linewidth]{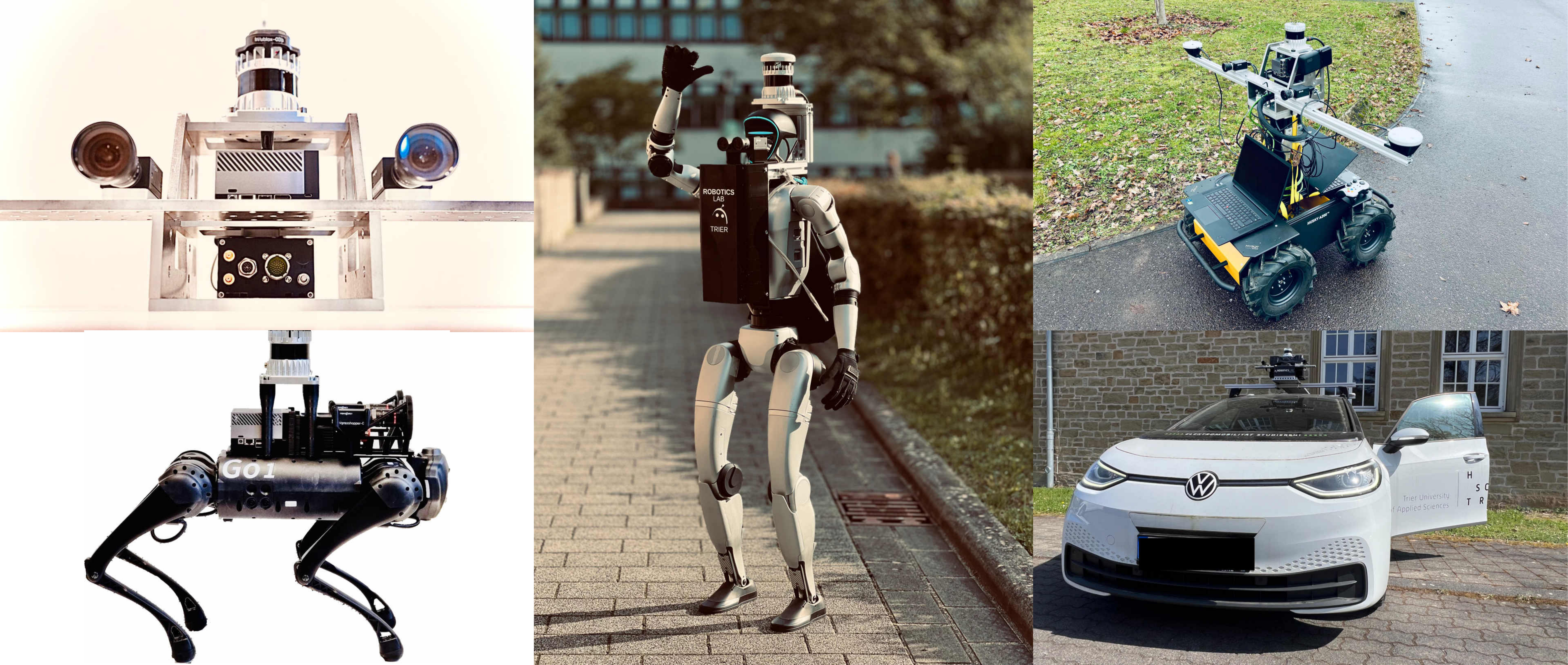}}
        \caption{
            The different mobile cognitive systems used within our laboratory,
            utilizing the same sensory system (top left) used for data acquisition
            in the present paper.
            Description and images modified from \cite{Schuler}.
        }
        \label{fig:fig1}
    \end{figure}

    Within our laboratory, we deploy various different mobile platforms including
    wheelchairs, humanoid, quadruped robots, wheeled vehicles as well as drones,
    with a selection displayed in Fig.~\ref{fig:fig1}.
    These different platforms are used in domains including indoor and outdoor
    campus as well as urban environments.
    For sensory systems, we utilize a flexible combination of multiple video
    cameras, LiDAR, INS and GNSS that can be adapted to our use cases.
    For our proposed pipeline, we assume a minimum of one video camera and
    INS/GNSS to be available.

    We want to highlight the versatility of our proposed pipeline within the
    context of our cognitive systems and domains, that is we make no assumptions
    about the system the pipeline is used with and the domains it is applied to,
    giving us full flexibility without any need to fine-tune the used models to
    a specific task or domain.
    As the edge device, we utilize either one or multiple NVIDIA Jetson AGX Orin,
    which host a custom C++ software stack developed within the laboratory to
    control these systems to handle the recording of all data and inference
    of Deep Learning applications as well as the pipeline implementation.
    Additionally, the pipeline itself can be used in a post-processing context
    on other local machines.
    For implementation details of the Deep Learning models specifically relevant to
    the pipeline refer to Sect.~\ref{subsec:experimental-setup}.

    \subsection{Automated Report Generation for Mobile Robotics}\label{subsec:automated-report-generation-for-mobile-robotics}

    For the generation of report documents for mobile agents, we focus on
    several distinct tasks important to the ability to interact with
    unstructured and unpredictable environments.
    This includes the detection and segmentation of objects, point cloud data
    of agents and objects of importance, the semantic description of sequences
    beyond the obvious and finally the relative and absolute localization of
    such agents, objects and descriptions within the world.

\begin{figure}[htbp]
        \centerline{\includegraphics[width=\linewidth]{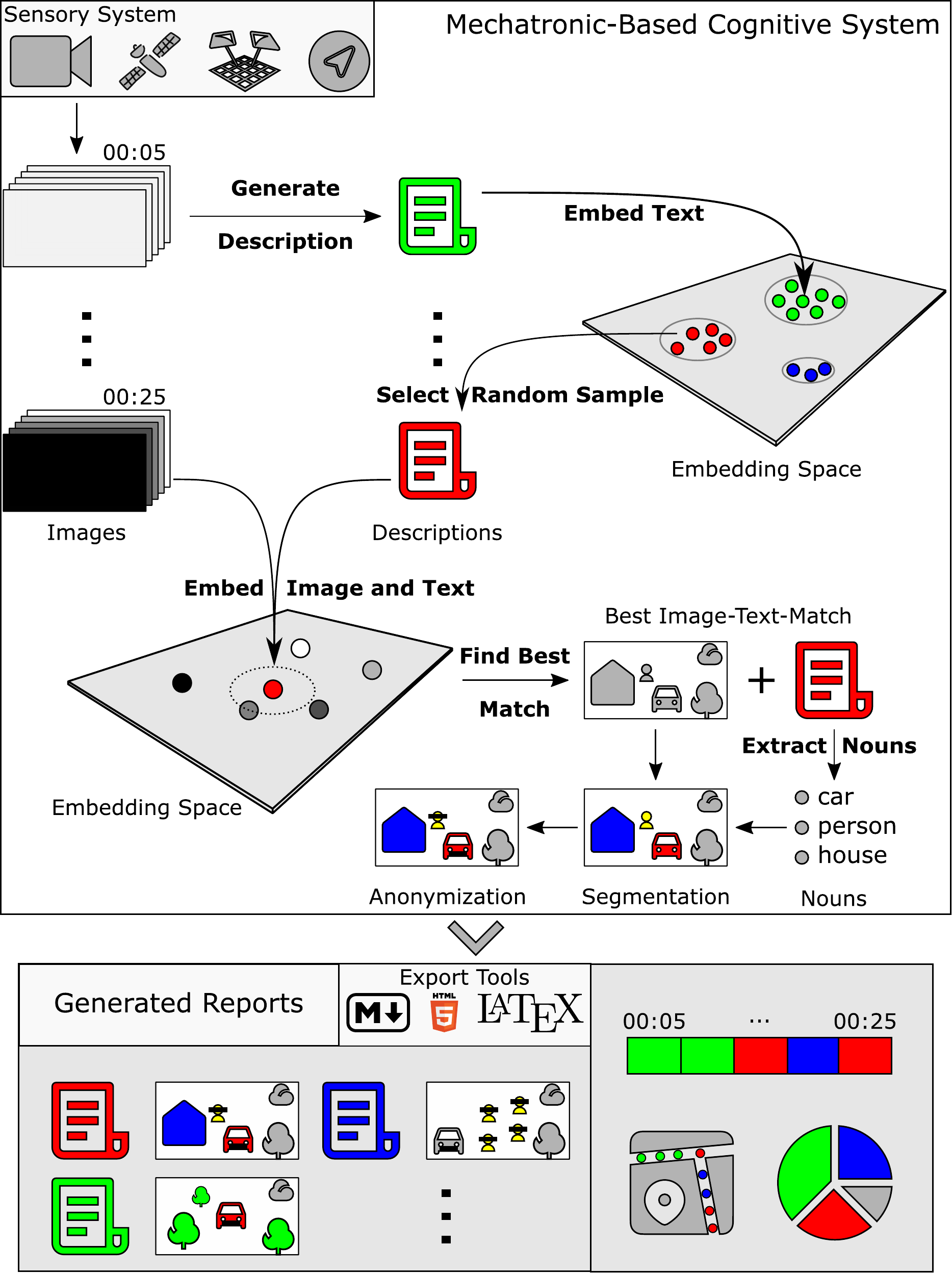}}
        \caption{The proposed pipeline for automated report generation for Mobile
        Robotics.
        We utilize a combination of VLM, text and image embedders like
        CLIP\cite{Radford2021}, NLP and zero-shot detection and segmentation models to
        generate reports of multi-modal data recorded by a variety of mobile cognitive
        systems, agnostic to the application domain.}
        \label{fig:fig2}
    \end{figure}

    To manage the large amount of data and make them accessible for human
    collaboration, our goal is to condense these information into digestible portions.
    We achieve this by clustering the generated descriptions by semantic
    similarity\cite{Yang2024} and by limiting the segmentation process to entities
    that are relevant to the described actions.
    Fig.~\ref{fig:fig2} provides an overview of the proposed architecture.

    First, the system generates a natural language description of a sequence of
    images in real time, using a local sVLM model.
    Note that we do not use all images within a sequence but sample them in an
    interval, for example one image per second, to enable real-time inference
    while retaining accuracy.
    These descriptions are not only used for report generation, but also during
    runtime.
    The generated descriptions and images are then saved for the final report
    generation.
    These textual descriptions are then embedded in a common embedding space
    using transformer models for textual embeddings\cite{Cao2024}.
    The resulting embedding vectors are clustered, forming groups of
    semantically similar descriptions.
    Next, a random description from each identified group is selected and
    embedded with its associated images using CLIP\cite{Radford2021} to a
    shared embedding space to calculate the best match between description and
    image candidates.
    We do this to exclude those images from the sequence that are not relevant
    to the generated description, e.g.\ in a batch of five images only the last
    three actual might contain the relevant agent mentioned in the description.
    Once the best image-text pair is found, we extract all nouns from the
    generated description using classical Natural Language Processing (NLP)
    via spaCy\cite{ExplosionAI}.
    These extracted nouns are used as text prompts for open-vocabulary
    detection and segmentation.
    The resulting images are finally anonymized using local Deep Learning
    models before being used in the report to preserve the privacy of all
    actively and passively involved agents.
    This is done for every identified cluster of descriptions, with the
    individual image-text pairs representing the cluster in the final document.

    Finally, we generate a report with the image-text pairs, the generated
    clusters, all generated descriptions and the localization of these grouped
    descriptions within the world.
    We assume no fixed output data representation format within the pipeline
    and thus are flexible to generate multiple reports in different formats,
    include Markdown, HTML and \LaTeX, all serving different purposes.
    Markdown documents might be used within documentations and repositories,
    also allowing for quick conversion into HTML documents, which contain
    interactive maps.
    These then can be easily displayed in web browsers or other applications.
    \LaTeX~finally generates highly customizable and structured professional
    PDF documents.

    \section{Experimental Evaluation}\label{sec:experimental-evaluation}

    During preliminary evaluation, we identified two critical points of failure
    for the proposed pipeline.
    First, the correctness and quality of the generated descriptions by the
    local VLM and second, the results of the clustering algorithm that groups
    these descriptions by Semantic Similarity.
    We already investigated the first aspect, that is the correctness of the
    generated textual descriptions by VLMs running on local, mobile platforms
    in a previous work\cite{Schuler}, and thus will focus on the second aspect
    for this evaluation.
    Note that we use the same dataset for both evaluations.

    \subsection{Experimental Setup}\label{subsec:experimental-setup}
    The proposed pipeline is implemented in Python 3.13 using
    PyTorch\cite{Paszke2019}, Transformers\cite{Wolf2019} and Sentence
    Transformers\cite{Reimers2019}.
    We chose Grounded DINO\cite{Liu2023} as the zero-shot object detection
    model, SAM\cite{Kirillov2023} as the zero-shot segmentation model and
    SmolVLM2\cite{Marafioti2025} as the sVLM.\@
    The communication between the C++ software stack and the Python modules is
    handled using LitServe\cite{LightningAI}.
    For the purpose of this work, we generate the reports via \LaTeX~as a PDF
    file as well as interactive HTML-based maps.
    For evaluation, we use a dataset consisting of 234 minutes of data,
    recorded in the German city of Trier by our laboratory.
    Examples from the dataset are given in Fig.~\ref{fig:fig4}.
    Within the dataset itself, we differentiate between three domains:
    \textit{Campus Indoor}, \textit{Campus Outdoor} and \textit{City}, with a
    split of 107 minutes, 74 minutes and 53 minutes respectively.
    We further split the dataset 20/80, with the first 20 \% being used for
    hyperparameter tuning and 80 \% for the final evaluation, keeping the ratio
    between domains.
    The individual sequences are split into five second clips that are fed into
    the VLM, generating textual descriptions of the scene, with the VLM
    sampling one frame per second to enable real-time inference on the edge
    device.
    The generated descriptions are then clustered by human annotators by
    similarity.
    We provide the instructions for the annotator in the supplementary
    materials.
    We do not evaluate the correctness of the generated descriptions itself and
    solely focus on the clustering of these descriptions, since the correctness of
    the descriptions in relation to the real world is of no concern to the quality
    of the semantic clustering itself.
    For a detailed study on the correctness of the generated description within
    our domains see our previous work\cite{Schuler}.

    For the quantitative evaluation, we compare the results of the automated
    semantic clustering using Sentence Transformers\cite{Reimers2019} to the manual
    clustered ground truth, reporting the Adjusted Rand Index
    (ARI)\cite{Steinley2004}, Normalized Mutual Information (NMI)\cite{Shannon1948}
    and Fowlkes-Mallows Index (FMI)\cite{Fowlkes1983}.
    We select the best hyperparameters per domain for the semantic clustering,
    using the highest ARI score on the first 20 \% of the data and report the
    discussed metrics on the final 80 \% used for evaluation.
    The hyperparameters optimized are the text embedding model as well as the
    metric and distance threshold used by the clustering algorithm.
    We provide the full data of the hyperparameter tuning process in the
    supplementary materials, see Sect.~\ref{sec:introduction}.
    Note that we do not fine-tune any of the Deep Learning models involved for a
    specific domain and only apply hyperparameter tuning to the clustering
    algorithm to better align them with our different domains and to evaluate if
    such a tuning is needed for specific domains to begin with.
    In addition to the quantitative evaluation, we also present qualitative
    results, including generated reports and findings that highlight successes and
    failures of the proposed pipeline.

    \subsection{Experimental Results}\label{subsec:experimental-results}

    The quantitative results as a measure of similarity between the ground truth
    clustering and the result of our pipeline are given in Tab.~\ref{tab:tab1}.
    A value closer to $1.0$ indicates higher similarity and for ARI and NMI,
    a value close to $0.0$ indicates a similarity close to chance level.
    The domain with the highest similarity between ground truth and semantic
    clustering is \textit{Campus Indoor} with an ARI of $0.620$, NMI of $0.793$ and an FMI
    of $0.721$.
    The domain \textit{City} show the lowest values in similarity to the ground truth with
    $0.476$, $0.654$ and $0.628$ respectively.

    For qualitative results, we provide example figures from an authentic report
    generated by the proposed pipeline, see Fig.~\ref{fig:fig3}.
    In addition, we provide multiple full reports in the supplementary material,
    see Sect.~\ref{sec:introduction}.

    \begin{table}[htbp]
        \caption{Results of the evaluation by domain.}
        \centerline{\includegraphics[width=\linewidth]{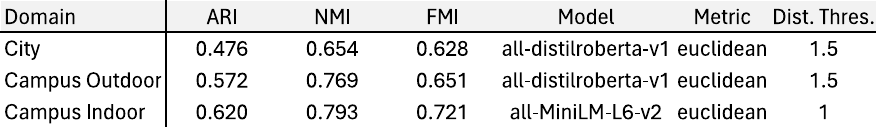}}
        \label{tab:tab1}
    \end{table}

    \begin{figure}[htbp]
        \centerline{\includegraphics[width=\linewidth]{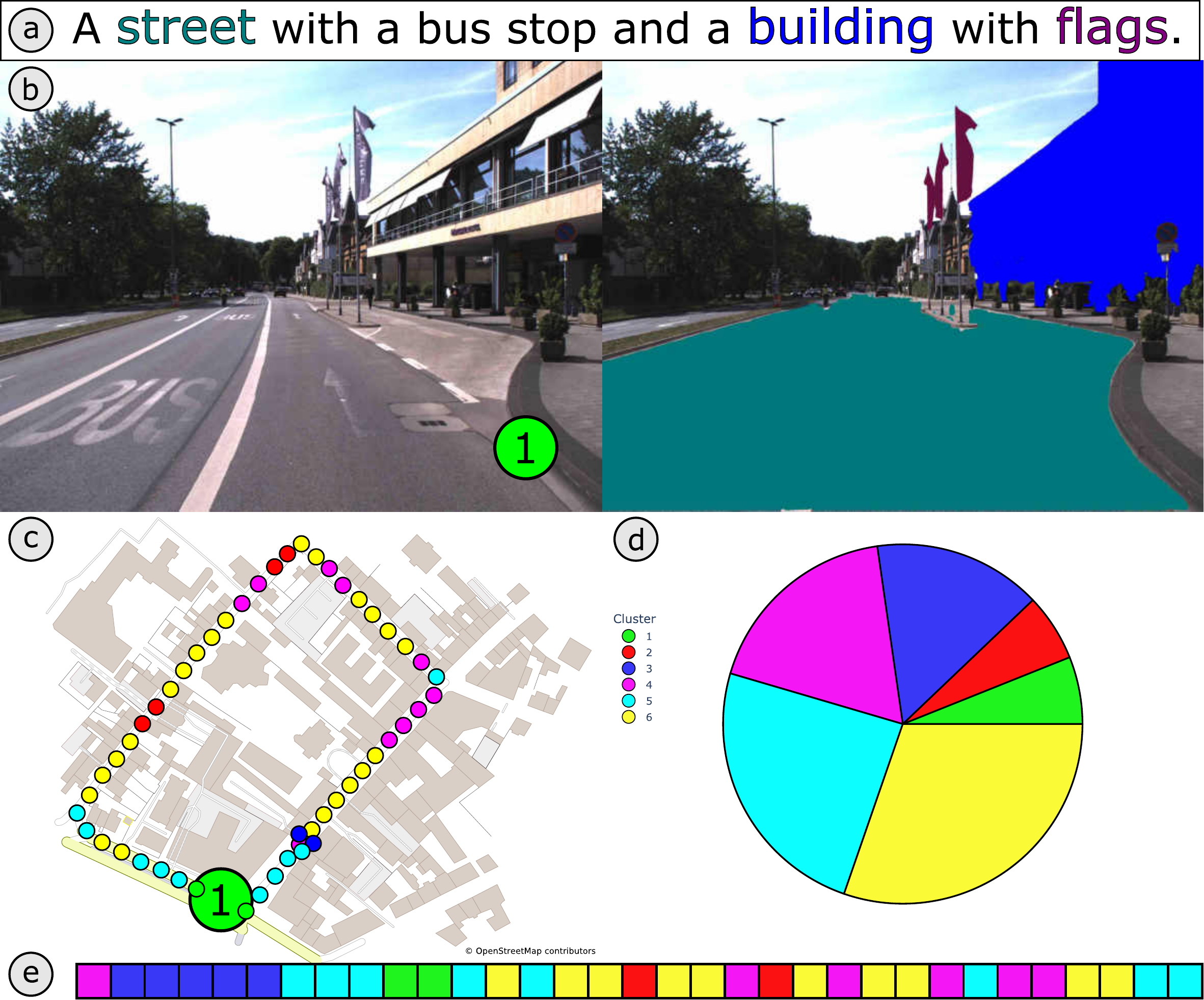}}
        \caption{
            Examples of an automatically generated report from
            a car drive through the German city of Trier.
            a) generated description,
            b) original image and segmentation results,
            c) location of individual descriptions colored by cluster,
            d) relative distribution of clusters and
            e) timeline of individual descriptions colored by cluster.
            Map modified from OpenStreetMap\cite{OSMF}.
        }
        \label{fig:fig3}
    \end{figure}

    \subsection{Discussion}\label{subsec:discussion}

    First, the discussion focuses on the quantitative evaluation, that is the
    measured similarity of the ground truth to the semantic clusters of the
    textual scene and action descriptions generated by a local VLM.\@
    We follow this up by discussing an authentic, generated report from our
    pipeline in greater detail.

    All three domains show values well above $0.0$ for all metrics
    respectively, displaying results of the clustering above the chance level
    at $0.0$ for the ARI and NMI scores.
    Looking at the individual domains, \textit{City} displays the lowest similarity to the
    ground truth across all three metrics used, while \textit{Campus Indoor} displays the
    highest.
    This is contrary to the findings in one of our previous works\cite{Schuler},
    where this domain had the highest correctness concerning the generated
    descriptions, with \textit{Campus Indoor} having the lowest.
    This fact might be explained by the difference in task at hand.
    The domain \textit{City} contains fewer radically different actions and scenes due to
    the relative monotony of urban traffic compared to highly dynamic indoor
    domains in a campus setting containing kitchens, offices, laboratories et
    cetera, which differences in descriptions often depend on small, hard to observe
    details.
    This facilitates the task of generating correct textual descriptions for the
    domain \textit{City}, but complicates the clustering of the generated descriptions.
    In some cases, the entire sequence might contain identical fragments in each
    description, e.g.\
    `A street with cars parked on the side and a few pedestrians walking on the sidewalk.',
    `A street with cars parked on the side and a few people walking on the sidewalk.' and
    `A street with cars parked on the side and a building in the background.'
    While we would consider the first two description to belong to the same
    cluster, one can argue that the third description belongs to a different
    cluster.
    The semantic clustering algorithm however might struggle to differentiate between
    this difference, since half of the sentence is identical for all three examples
    provided.
    This can be tweaked using hyperparameters, but that might lead to other
    descriptions not being clustered correctly together anymore.

    The results of the hyperparameter optimization show identical optimal
    parameters for the domains \textit{City} and \textit{Campus Outdoor} and we want to note that
    for the domain \textit{Campus Indoor}, the same hyperparameter configuration is in the
    top five combinations as well.
    This indicates to us that the semantic clustering using transformers is stable
    enough to use identical tuned hyperparameters for all domains while achieving
    satisfactory clustering.
    Since our approach aims to be agnostic to its domain, environment and
    cognitive systems used, this is a highly desirable property to achieve,
    especially since manual evaluation using human annotators is labor-intensive
    and subjective, thus prone to biases\cite{Gautam2024}.
    While we aim to reduce this subjectivity by guiding the annotation
    process (see supplementary materials), this challenge remains.

    We want to discuss qualitative results of our evaluation, that is authentic
    results from the generated reports.
    As displayed in Fig.~\ref{fig:fig3}, the clustering and localization within the
    report gives us insights into our domain and environment.
    The magenta cluster describes actions exemplified by `A cyclist is riding down
    a city street.' and is clustered in the side streets of this particular drive
    (to the top of the map), while the turquoise cluster, exemplified by `A
    street scene with cars driving down the road.' mainly occurs on the main
    streets within the city (to the bottom of the map).
    The green cluster, exemplified by `A street with a bus stop and a building with
    flags.', highlights a bus lane and close-by hotel with flags in front of
    it located at the city centre.
    Finally, the yellow cluster, exemplified by `A street with cars and people
    walking on the sidewalk.', is a more generic cluster which contains
    descriptions that are frequent both on the side and main streets.
    These results can not only be used for manual reviews by laymen, but also to
    augment semantic maps, highlighting spatio-temporal similarities and frequency
    of descriptions connected to certain streets and locations.
    The semantically guided zero-shot segmentation process further highlights the
    area of interest and thus location of objects related to the description.
    Since the absolute pose w.r.t.\ various geodetic data and time of the mobile
    cognitive system is known, this allows us to exactly localize these objects of
    interest within the world of the mobile system.
    In addition, we are able to differentiate between active agents, e.g.\ humans
    and vehicles, and passive objects we assume to not move, e.g.\ buildings,
    enabling us to build dynamic semantic maps comprised of description and object
    heatmaps.

    \section{Conclusion and Future Work}\label{sec:conclusion-and-future-work}

    In this work, we presented a privacy-preserving, local pipeline for automated
    report generation for mechatronic-based cognitive systems.
    We demonstrated that our system is able to generate semantic clusters similar
    to a human annotator, with the resulting report offering valuable insights
    into the environment of the mobile cognitive system.

    For future work, we aim to add more sensor information provided by our platform
    specific to the task at hand, in particular by incorporating AI responsible for
    decision-making on our mobile cognitive system.
    Further, we want to expand the internal data representation to incorporate
    scene graphs and additional semantic information, especially to increase the
    explainability and internal consistency within our pipeline, expanding on the
    topic of automated semantic map generation and world models\cite{Guan2024}.
    We also want to utilize the aspect of collective robotics in building shared,
    detailed maps, leveraging consistent internal world representations.

    \section*{Appendix}\label{sec:appendix}

\begin{figure}[htbp]
        \centerline{\includegraphics[width=\linewidth]{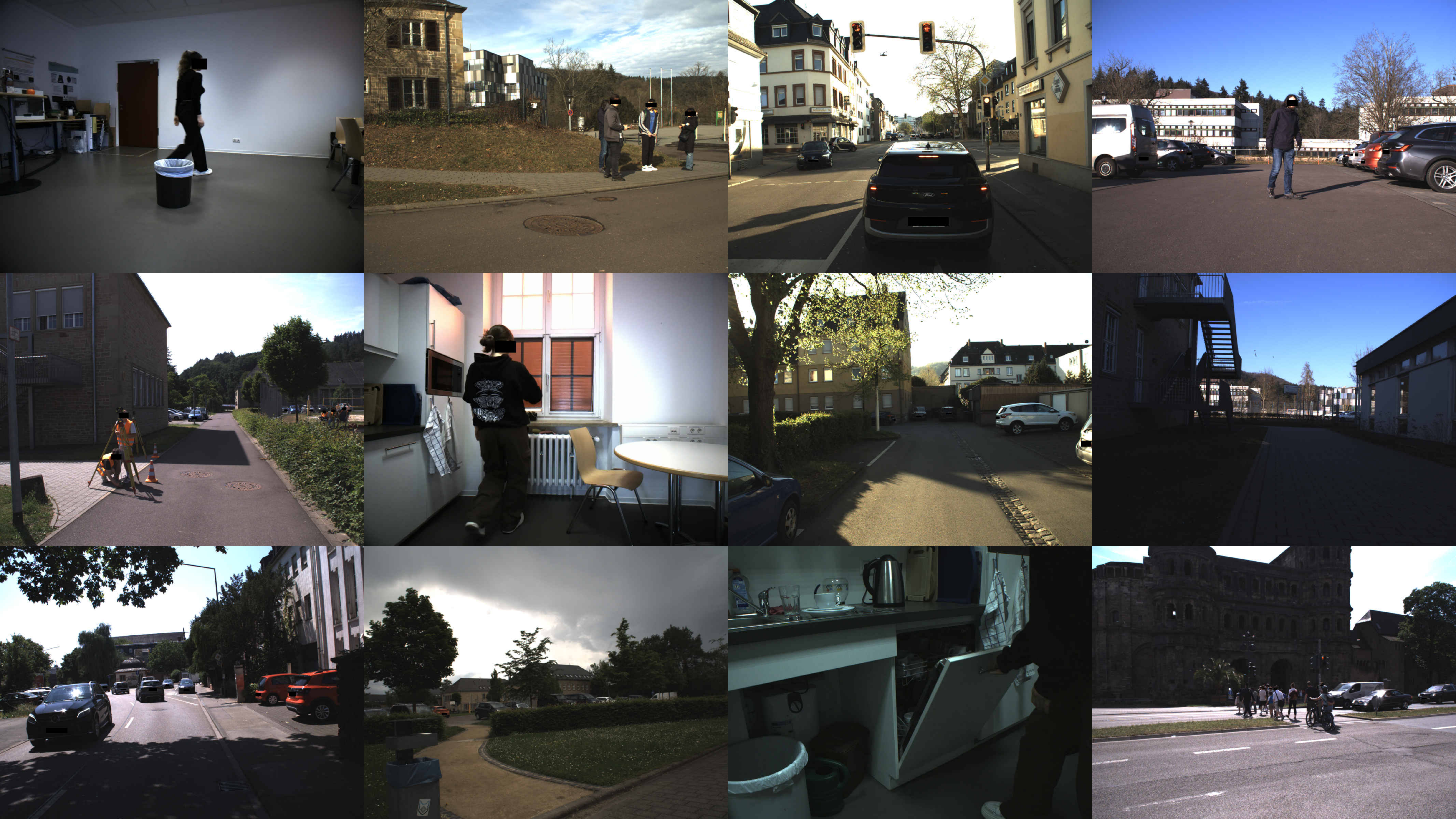}}
        \caption{Examples from the dataset used in the evaluation of this paper.
        The dataset contains various domains, challenging lighting conditions,
            crowded scenes and complex scenarios.
            The dataset was captured by different mobile cognitive systems within
            our laboratory.
            Images taken from \cite{Schuler}.
        }
        \label{fig:fig4}
    \end{figure}

    Fig.~\ref{fig:fig4} gives selected examples of the dataset used for evaluation.
    The dataset was recorded in the German city of Trier, containing three
    domains: \textit{City}, \textit{Campus Outdoor} and \textit{Campus Indoor}.
    Note that we anonymize the images only for the purpose of this publication, the
    models and mobile cognitive systems utilize the raw data instead.

\end{document}